\documentclass[11pt,a4paper]{article}
\usepackage[hyperref]{acl2019}

\usepackage{times}
\usepackage{latexsym}
\usepackage{algorithm}
\usepackage{setspace}
\usepackage{amsfonts}       % blackboard math symbols
\usepackage{amsmath}
\usepackage{algcompatible}
\usepackage{mathtools}
\usepackage{url}
\usepackage{nicefrac}       % compact symbols for 1/2, etc.
\usepackage{multirow}
\usepackage{bm}
\newcommand{\vect}[1]{\bm{#1}}
\newcommand{\matr}[1]{\bm{#1}}

\newcommand{\va}[0]{\vect{a}}
\newcommand{\vb}[0]{\vect{b}}
\newcommand{\vc}[0]{\vect{c}}

\newcommand{\vh}[0]{\vect{h}}
\newcommand{\vv}[0]{\vect{v}}
\newcommand{\vx}[0]{\vect{x}}

\newcommand{\vs}[0]{\vect{s}}

\newcommand{\vi}[0]{\vect{i}}
\newcommand{\vo}[0]{\vect{o}}
\newcommand{\vy}[0]{\vect{y}}

\newcommand{\vp}[0]{\vect{p}}
\newcommand{\vtheta}[0]{\vect{\theta}}

\newcommand{\mW}[0]{\matr{W}}

\newcommand{\mS}{\matr{S}}

\aclfinalcopy

\title{Adversarial reconstruction for Multi-modal Machine Translation}

 \author{Jean-Benoit Delbrouck \and St\'ephane Dupont \\
         TCTS Lab, University of Mons, Belgium\\
          \{jean-benoit.delbrouck, stephane.dupont\}@umons.ac.be}

\date{}

\begin{document}
\maketitle

\begin{abstract} Even with the growing interest in problems at the intersection of Computer Vision and Natural Language, grounding (i.e. identifying) the components of a structured description in an image still remains a challenging task. This contribution aims to propose a model which learns grounding by reconstructing the visual features for the Multi-modal translation task. Previous works have partially investigated standard approaches such as regression methods to approximate the reconstruction of a visual input. In this paper, we propose a different and novel approach which learns grounding by adversarial feedback. To do so, we modulate our network following the recent promising adversarial architectures and evaluate how the adversarial response from a visual reconstruction as an auxiliary task helps the model in its learning. We report the highest scores in term of BLEU and METEOR metrics on the different datasets.
\end{abstract}

\section{Introduction}
Problems combining vision and natural language processing are viewed as a difficult task. It requires to grasp and express low to high-level aspects of local and global areas in an image as well as their relationships. Visual attention-based neural decoder models \cite{pmlr-v37-xuc15,KarpathyL15} have been widely adopted to solve such tasks. The attention focuses only on part of an image and integrates this spatial information into the multi-modal model pipeline. The model, that usually consists of a Recurrent Neural Network (RNN), encodes the linguistic inputs and is trained to modulate, merge and use both visual and linguistic information in order to maximize a task score. For instance, in Multi-modal Machine Translation (MMT), the model is required to translate an image description to another language.\\

The integration of visual input in MMT has always been the primary focus of the different researches in the field. Regional and global features have first been investigated \cite{huang2016attention}, then convolutional features of higher dimensions (such as the res4f layer from ResNet) \cite{calixto2017doubly, delbrouck2017multimodal} were used because they carry more visual information. Recently, \citet{caglayan2017lium} found that light architectures with fewer parameters are more suitable for the learning of the MMT task. Because of the limited number of training parameters, global features must be used. A trade-off arises : models with bigger attention mechanism could take advantage of richer visual input but the addition of training parameters seems to impair the translation quality. \\

To tackle this problem, we decide to take a state-of-the-art MMT model and add a conditional generator whose aim is to reconstruct the global visual input used during the translating process using only the model terminal state. We also want this reconstruction to be evaluated adversarially. This approach has four purposes :

\begin{itemize}
    \item We constrain the model to closely represent the semantic meaning of the sentence by reconstructing the visual input. We believe it  would  ground  the  visual information  into  the  training  process and enable better generalization;
    \item We leave the whole translation model pipeline unchanged, no learning parameters are added for translation. The generator module is trained end-to-end during training but is unused during inference;
    \item Because we use light global features, the reconstruction process is very fast and require few learning parameters;
    \item By using an adversarial approach, we want our generator to approximate the true data distribution of images.  We believe that the propagation of generator's gradient back into the translation model would enable better generalization for unseen images on the different test-sets.
\end{itemize}

This reconstruction problem has two parts. First, we add the reconstruction module on top of our primary MMT task and investigate the different architecture for the generator. Secondly, we treat the reconstruction as an adversarial problem. We modulate our network following recent promising adversarial architectures and evaluate how the adversarial response helps the translation pipeline in its learning. We prove their efficiency by showing strong generalization the different MMT test-sets.

\section{Related work}
In the modality reconstruction field, the closest work related to ours is the one of \cite{rohrbach2016grounding} who proposes an approach which can learn to visually localize phrases relying on phrases associated with bounding boxes in an image. Nevertheless, our works differ in two ways. First, the reconstruction is linguistic. They aim to reconstruct the sentence from a visual attention. Secondly, their visual data are annotated with bounding boxes representing linguistic information while our approach doesnt require any preprocessing. \\

When reconstructing its input, a model can be seen as an auto-encoder \cite{hinton2006reducing} which aims to compress or encode with model $Q(z|v)$ a modality $v$ into a representation $z$  and then decode (or reconstruct) from z an approximation $v^\prime$ with decoder $G(v|z)$. The difference lies in that our latent variable $z$ (or compressed representation) is the final representation of a MMT model. Input $v$ is modulated in by the multi-modal model before being decoded (or reconstructed). Because our latent variable $z$ will be adversarially evaluated, our model is also close to an adversarial auto-encoders (AAE) \cite{aae}. \\

Adversarial approaches for multimodal-tasks have been investigated in image-captioning \cite{feng2018unsupervised} or visual question answering \cite{ilievski2017generative}. In those works, the task goal is fully adversarial which differs from our approach. Our translation model is still a classification task and uses the widely adopted negative log likelihood loss. Only the reconstruction module is treated as adversarial.\\

Finally, reconstruction (or imagination as called in the author's paper) has been investigated with regression techniques \citet{elliott2017imagination}. A major difference, besides our adversarial approaches, is their choice to not use any visual information during inference. The image is only used as training input for the reconstruction module, not the translation module. We believe it could penalize the model to do so if the information for translation really is in the image. As previously stated, using visual input for translation might impair overall translation quality but we force our model to use a visual attention during inference as it is the very foundation of the multimodal translation task.

\section{Background}
In this section, we describe the concepts involved in our experiments. We start by describing how visual reconstruction as an auxiliary task is built on top of our MMT model. We then explain the two adversarial settings used to  involved in our experiments: a generative adversarial network and an adversarial auto-encoder.

\subsection{Visual reconstruction} \label{viz_rec}

We denote the MMT model $\mathnormal{Q}$ and its inputs $\mathnormal{x}$ and $\mathnormal{v}$ for the linguistic and visual data respectively. The model learns to output the translation $\mathnormal{y}$ of $x$ as formulated hereafter:
\begin{align}
\mathnormal{y}, {h_T} &= \mathnormal{Q}(\mathnormal{x},\mathnormal{v})\label{eq:model}
\end{align}
where ${h_T}$ is defined as the model's $Q$ final state (or last hidden state). A generator $G$ takes as input ${h_T}$ and approximates a visual reconstruction $\mathnormal{{v^\prime}}$:
 \begin{align}
\mathnormal{v^\prime} &= \mathnormal{G}({h_T})\label{eq:generator}
\end{align}
From equation \ref{eq:model} and \ref{eq:generator}, we compute the total loss $\mathcal{L}_\mathnormal{MMT}$ of model $Q$ and generator $G$ :
\begin{equation}
\mathcal{L}_\mathnormal{MMT} = \overbrace{\mathcal{L}_\mathnormal{Q}(y,x)}^{\text{translation pipeline}}+  \ \lambda_r \ \: \overbrace{\mathcal{L}_\mathnormal{R} (\mathnormal{v^\prime},\mathnormal{v})}^{\text{reconstruction pipeline}}
\end{equation} 
Factor $\lambda_r$ indicates the weight of the reconstruction loss.\\

Notation used in this sub-section \ref{viz_rec} are matched in the following sub-sections \ref{gan_sec} and \ref{aae_sec} for clarity.

%GAAAAAAAAAAAAAAAAAAAAAAAAAAAAAAAAAAN

\subsection{Generative adversarial network (GAN)} \label{gan_sec}
A generative adversarial network \cite{gan} is a model whose main focus is to generate new data based on source data. It is made of two networks: the generator $G$ that constructs synthetic data from noise samples $z$ and the discriminator $D$ that distinguishes generated samples from the generator or from the true data-set distribution. 
Intuitively, one can say that the goal of the generator is to fool the discriminator by synthesizing data close to the data distribution. This leads to a competition between both networks called the min-max objective:

\begin{equation}
\begin{aligned}
    \min_{G} \max_{D} \mathbb{E}_{v \sim \mathbb{P}_{true}}&[\log (D(v))] + \\ & \mathbb{E}_{v^\prime \sim \mathbb{P}_{generated}}[\log (1 - D(v^\prime))] 
\end{aligned}
\end{equation}

where $v$ is an example from the true data and $v^\prime = G(z)$ a sample from the Generator and variable $z$ is Gaussian noise.\\

To stabilize training and tackle the vanishing gradient problem, \citet{wgangp} introduce a gradient penalty in the objective : 
\begin{equation}
\begin{aligned}
    \mathbb{E}_{v \sim \mathbb{P}_{true}}[D(v)] &+ \mathbb{E}_{v^\prime \sim \mathbb{P}_{generated}}[1 - D(G(z))] \\ &+  \lambda_{gp} \; \mathbb{E}_{ \hat v \sim \mathbb{P}_{\hat v}}[(\Vert {\nabla_{\hat v}D(\hat v)}\Vert_2-1)^2] \label{eq:wgan_gp}
\end{aligned}
\end{equation}

with $\hat v = \epsilon v + (1-\epsilon) v^\prime$ and where $\epsilon$ is a random number sampled from the uniform distribution $U[0,1]$ and $\lambda_{gp}$ is the penalty factor. This method produces more stable gradient and the critic can match more complex distribution.

This equation refers to the wasserstein GAN (WGAN, \citet{wgangp}) with gradient penalty that will be used in our experiments at section \ref{expe}.

%AAAAAAAAAAAAAAAAAAAAAAAAAAAAAAAAAAAAAAAAAAAAAAAAAAE

\subsection{Adversarial Auto-encoders (AAE)} \label{aae_sec}

Auto-encoders are made of two parts : an encoder $Q$ receives the input $v$ and creates a latent or hidden representation $h$ of it, and the generator $G$ takes this intermediate representation and tries to reconstruct the input as $v^\prime$. A common loss is to use the mean square error between the input and reconstructed inputs.

\begin{equation}
    L_R (, v') = ||v - v'||^2
\end{equation}

Variational autoencoders impose a constraint on how to construct the hidden representation. The encoder can not use the entire latent space freely but has to restrict the hidden codes $h$ produced to be likely under the prior distribution $p(v)$. This can be seen as a type of regularization on the amount of information that can be stored in the latent code. The benefit of this relies on the fact that now we can use the system as a generative model. To create a new sample that comes from the data distribution $p(v)$, we sample from $p(h)$ and run this sample through the generator. In order to enforce this property a second term is added to the loss function in the form of a Kullback-Liebler (KL) divergence between the two distributions : 

\begin{equation}
L(v, v') = L_R (v, v') + KL(Q(h|v)||p(h))
\end{equation}
 
where $Q(h|v)$ is the encoder of our network and $p(h)$ is the prior distribution imposed on the latent code.\\

Adversarial autoencoders \cite{aae}  avoid using the KL divergence by using adversarial learning. In this architecture, a new discriminative network $D$ is trained to predict whether a sample comes from the latent code of the generator $Q(h|v)$ or from the prior distribution imposed on the latent code $p(h)$. The loss of the encoder is now composed by the reconstruction loss plus the loss given by the discriminator network. \\

We can now use the loss incurred by the encoder of the adversarial network instead of a KL divergence for it to learn how to produce samples according to the distribution $p(h)$. The loss of the discriminator $D$ is :

\begin{equation}
 L_D = -\log(D(h')) + \log(1 - D(h))
\end{equation}
where $h$ is generated by the encoder and $h^\prime$ is a sample from the true prior (usually a gaussian distribution). Following the mix-max game, the loss of the encoder $Q$ is :
 \begin{equation}
 L_Q = -\log(D(z))
\end{equation}
As seen in the previous sub-section, we can make this AAE wasserstein (WAAE, \citet{waae}) by using the Wasserstein distance between the two probability distributions and by introducing a regularizer penalizing discrepancy between prior distribution and distribution induced by the encoder.

\section{MMT Experiments} \label{expe}

In this section, we describe the two visual reconstruction experiments on model $Q$ evaluated in section \ref{results}.

\subsection{$G$-WGAN}

\begin{figure}[ht!]
	\centering
	\includegraphics[scale=0.26]{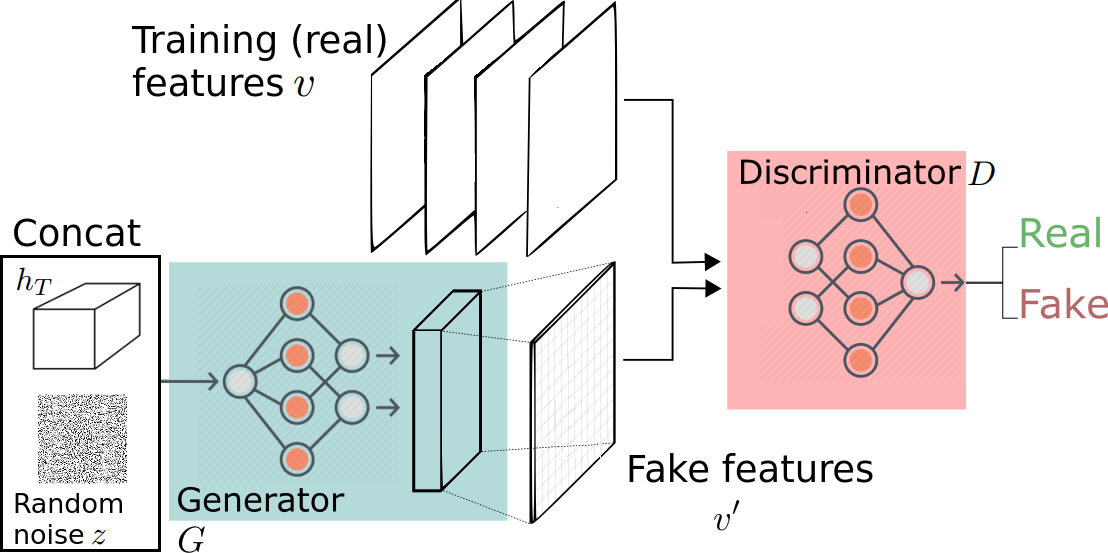}
	\caption{Training flow of $G$-WGAN. Model $Q$ omitted for clarity.}
	    \label{gwgan}
\end{figure}

In the original algorithm, G receive $z$ as input and is usually a sample from Gaussian noise. In the case of MMT, noise $z$ will be concatenated with the model $Q$'s last hidden state $h_T$ so that the generator reconstruct the features according to the translated sentence. Generator $G$ then becomes a conditional generative network \cite{cgan} and outputs the reconstructed features ${v^\prime} = G([z,h])$. This reconstruction will be evaluated by discriminator $D$. This settings is illustrated in figure \ref{gwgan}. The goal of noise is to make the generator non-deterministic so that it is harder the for model $D$ to discriminate between the real and the fake sample. Stochasticity can be induced by dropout as well \cite{isola2017image} and will be used in our model. The full procedure can be found in Algorithm \ref{alg-g}.

\subsection{$Q$-WAAE}
\begin{figure}[h!]
	\centering
	\includegraphics[scale=0.30]{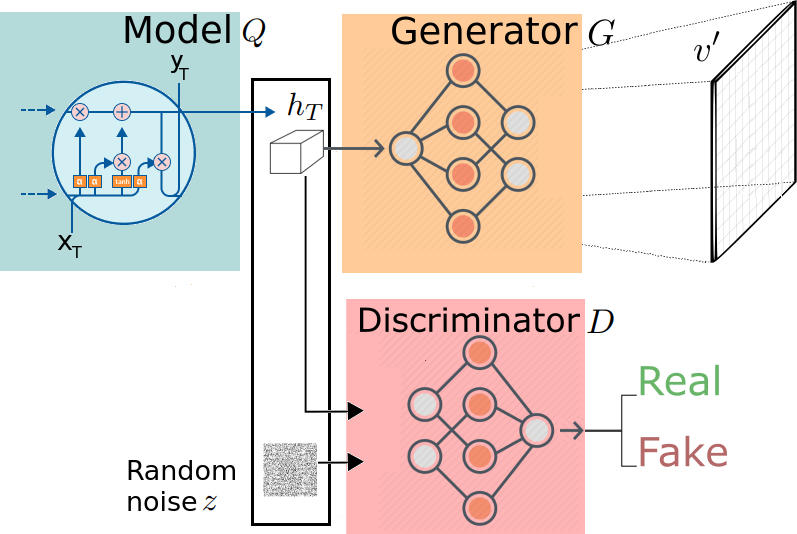}
	\caption{Training flow of $Q$-WAAE. The last hidden state $h_T$ is the input for decoder $P$}
     \label{fig_waae}
\end{figure}

In this experiment, the encoder $Q$ is actually the multi-modal translation model $Q$. The latent variable $h$ is seen as the last hidden state $h_T$ of the model $Q$. $D$ has to discriminate between the latent code $h_T$ or the "real" latent code $h^\prime$ sampled from a Gaussian distribution. Along the adversarial loss, a generator $G$ reconstruct the features $v^\prime$ with input $h_T$. The figure \ref{fig_waae} depicts the reconstruction. The full procedure can be found in Algorithm \ref{alg-q}.

\section{Settings}
In this section, we describe the model $Q$ and the data-set used.
\subsection{Training}
To be consistent with the state-of-the-art, we follow the settings that are used in the previous works we compare our model to in the result section.  The full description of the model $Q$ can be found in appendix \ref{app:modelq}. RNN layer size, attention size, dropout, model ensembling and training settings are left unchanged for a fair comparison.  \\ 

We train jointly $Q$ and $G$ with Adam optimizer \cite{kingma2014method} with the learning rate 4e-4 and gradient clipping is set to 1. The visual input $v$ used are the images features from the last pooling layer (pool5) of the ResNet-50 \cite{he2016deep} and are of dimension $2048 \times 1$. We use a batch-size of 32. For both task, we stop training if the task score doesn't improve for more than 5 epochs. Model reported are ensembling of 5 models. \\

Finally, the gradient penalty $\lambda_{gp}$ is set to 10 for all experiments. For $Q$-WAAE, the $\lambda_{critic}$ coefficient is set to 5. The adversarial and reconstruction coefficients $\lambda_a$ and $\lambda_r$ are detailed in the results section \ref{results}. The discriminator $D$ is trained with adam with learning rate of 2e-4, $\beta_1$ = 0.5 and $\beta_2$ = 0.9. The architecture of $G$ and $D$ is available in Appendix \ref{app:spectral}. We found out that the use spectral normalization \cite{miyato2018spectral} and batch normalization didn't improve the translation scores.

   \begin{algorithm}[t]
  \setstretch{1.1} 
  \caption{$G$-WGAN : Wasserstein GAN with gradient penality}
   \textbf{Require:}  Adversarial coefficient $\lambda_a > 0$, gradient penalty coefficient $\lambda_{gp} = 10$, the number of $D$ iterations per $G$ iteration $\lambda_{critic} = 5$
  \begin{algorithmic}[]
   \State Initialize the parameters $\theta$ of the MMT model $Q$, generator $G$ and features discriminator $D$.
    \State \textbf{while} Q not converged \textbf{do}
    \State \quad Sample $x,v$ from the training set
    \State \quad Output translations $y$ from $Q(x,v)$
    \State \quad Get last states $h_T$ from $Q$
    \State \quad \textbf{for} $t = 1, ..., \lambda_{critic}$ \textbf{do}
    \State \quad \quad Sample noise $z$ from $\mathcal{N}(0,1)$
    \State \quad \quad Sample random number $\epsilon$ from $U[0,1]$
    \State \quad \quad $v^\prime \leftarrow G([z,h_T])$
    \State \quad \quad $\hat v \leftarrow v \epsilon +  v^\prime(1-\epsilon) $
    \State \quad \quad Update $D_\theta$ by ascending:
    \quad \begin{equation*}
        \begin{multlined}
        D (v) + (1 - D({v^\prime}))\\ + 
        \lambda_{gp} \; (\Vert {\nabla_{\hat v}D(\hat v})\Vert_2-1)^2
        \end{multlined}
    \end{equation*}
    
    \State \quad Update $G_\theta$ and $Q_\theta$ by descending the adversarial loss $\mathcal{L}_R$:
    \begin{equation*}
          \lambda_a \; D(v^\prime)
    \end{equation*}
    \State \quad Update $Q_\theta$ by descending translation loss $\mathcal{L}_\mathnormal{Q}(x,y)$

  \end{algorithmic}
  \label{alg-g}
\end{algorithm}

\begin{algorithm}[t]
  \setstretch{1.1} 
  \caption{$Q$-WAAE : Wasserstein Auto-Encoder with gradient penalty}
  \label{fig:algo}
  \textbf{Require:} Adversarial coefficient $\lambda_a > 0$, reconstruction coefficient $\lambda_r > 0$, gradient penalty coefficient $\lambda_{gp} = 10$
  \begin{algorithmic}[]
      \State Initialize the parameters $\theta$ of the MMT model $Q$, generator $G$ and latent discriminator $D$. Use mean square error as $c$.
    \State \textbf{while} $Q$ not converged \textbf{do}
    \State \quad Sample $x,v$ from the training set
    \State \quad Output translations $y$ from $Q(x,v)$
    \State \quad Get last states $h_T$ from $Q$
    
    \State \quad Sample "true" state $h^\prime$ from $\mathcal{N}(0,1)$
    \State \quad Sample random number $\epsilon$ from $U[0,1]$
    \State \quad $\hat h \leftarrow h_T \epsilon +  h^\prime(1-\epsilon) $
    \State \quad Update $D_\theta$ by ascending:
    \begin{equation*}
        \begin{multlined}
        D (h^\prime) + (1 - D({h_T}))\\ + 
        \lambda_{gp} \; (\Vert {\nabla_{\hat h}D(\hat h})\Vert_2-1)^2
        \end{multlined}
    \end{equation*}
    \State \quad Update $G_\theta$ and $Q_\theta$ by descending reconstruction and adversarial loss $\mathcal{L}_R$:
    \begin{equation*}
         \lambda_r \; c(v,G(h_T)) - \lambda_a \; \log(D (h_T))
    \end{equation*}
    \State \quad Update $Q_\theta$ by descending translation loss $\mathcal{L}_\mathnormal{Q}(x,y)$

  \end{algorithmic}
    \label{alg-q}
\end{algorithm}

\subsection{Dataset}
We use the Multi30K dataset \citep{elliott-EtAl:2016:VL16}. For each image, one of the English descriptions was selected and manually translated into German by a professional translator. As training and development data, 29,000 and 1,014 triples are used respectively. We use the three available test sets to score our models. The Flickr Test2016 and the Flickr Test2017 set contain 1000 image-caption pairs and the ambiguous MSCOCO test set 461 pairs. Recently, a fourth dataset, the Flickr Test2018 set, is used for the online competition on codalab \footnote{{https://competitions.codalab.
org/competitions/19917\#results}}. It consists of 1,071 sentences is released without the German and French gold translations.

\section{Results} \label{results}
We now report the results for the different two configurations introduced in section \ref{expe} on the Multi-modal Machine Translation (MMT) task. All experiments reported were run on a single NVIDIA GTX 1080 GPU.

\begin{table*}[t]
\centering
\begin{tabular}{llllllll}
			\multicolumn{1}{c}{\bf Test sets}  &\multicolumn{2}{c}{\textbf{Test 2016 Flickr}} &\multicolumn{2}{c}{\textbf{Test 2017 Flickr}} 
			\\ \hline \\
			&\multicolumn{1}{l}{\bf 	BLEU}   &\multicolumn{1}{l}{\bf 	METEOR} 
			&\multicolumn{1}{l}{\bf 	BLEU}   &\multicolumn{1}{l}{\bf 	METEOR} \\
           
			\\ 
            FAA\shortcite{caglayan-EtAl:2018:WMT} & - & - & 31.60 & 52.50 \\
            DeepGru\shortcite{delbrouck2018umons} & 40.34 & 59.58 & 32.57 & 53.60 \\
            Baseline & 40.00 & 59.20 & 32.20 & 53.10\\
            $G$-WGAN  & 40.38 \footnotesize{+0.38} &  60.03 \footnotesize{+0.83}  & 33.70 \footnotesize{+1.50}& 54.50 \footnotesize{+1.40}\\
            $Q$-WAAE & 40.66 \footnotesize{+0.66} &  60.06 \footnotesize{+0.86}  & 34.06  \footnotesize{+1.86}& 54.94 \footnotesize{+1.84}\\

            \\
            \multicolumn{1}{c}{\bf Test sets}  &\multicolumn{2}{c}{\textbf{COCO-ambiguous}} &\multicolumn{2}{c}{\textbf{Test 2018 Flickr}} 
			\\ \hline \\
			FAA\shortcite{caglayan-EtAl:2018:WMT} & - & - & 31.39 & 51.43 \\
            DeepGru\shortcite{delbrouck2018umons} & 29.21 & 49.45 & 31.10  & 51.64 \\
			Baseline & 28.50 & 48.80 & - &-\\
            $G$-WGAN & 31.08 \footnotesize{+2.58} &   50.43 \footnotesize{+1.63}  & 31.80 \footnotesize{} & 52.15 \footnotesize{}\\
            $Q$-WAAE & 31.41 \footnotesize{+2.91} & 50.95 \footnotesize{+2.15}  & \textbf{31.91} \footnotesize{} &  \textbf{52.37} \footnotesize{}\\
                        \hline \\

\end{tabular}
\caption{Results on the en$\xrightarrow{}$de MMT task. Test 2018 results (anonymized) can be checked on the official leaderboard (\url{https://competitions.codalab.org/competitions/19917\#results}) in the "german" tab. Score differences are computed against the baseline.}
\end{table*}

%%% ajouter unclear si il faut adversarial, moduler h_t ou le cnn comme jbdelbrouck le sympa
%matrice confusion
%montrer le role de de la taile du bruit

\subsection{Quantity evaluation}
First and foremost, we notice that the most successful model is $Q$-WAAE as it marginally surpasses the baseline and previous works in every dataset. It is also the best official reported score as constrained submission (only data provided by the challenge) of the test 2018 data-set. The submission surpasses the previous best METEOR score from DeepGru by 0.73 METEOR and the previous best BLEU score from FAA by 0.52 points.  More importantly, the $Q$-WAAE model significantly improves the SOTA on the COCO-ambiguous data-set, a test-set that has been specifically designed to include 56 unique ambiguous verbs in 461 descriptions ($+2.91$ BLEU and $+1.63$ METEOR). \\

\begin{table}[!h]
  \centering
\def\arraystretch{1.5}
\begin{tabular}{cc|cccc}
\multicolumn{2}{c}{}
            &   \multicolumn{3}{c}{$\lambda_r$} \\
    &       &   0.2 &   0.5   &  0.8           \\ 
    \cline{2-5}
\multirow{3}{*}{\rotatebox[origin=c]{90}{$\lambda_a$}}
    & 0.2   & \textbf{50.95}   & 50.08    & 49.33              \\
    & 0.5    & 49.79   & 49.62   & 49.16               \\ 
        & 0.8    & 49.70    & 49.16    & 48.02              \\ 

    \cline{2-5}
    \end{tabular}
\caption{$Q$-WAAE : Impact on the METEOR metric of the reconstruction and adversarial loss coefficient on the ambiguous COCO data-set}
\label{waaelambda}
\end{table} 

To try and get the best results on the $Q$-WAAE, we mixed different combinations of the coefficient factors on the adversarial and reconstruction loss as shown in table \ref{waaelambda}. The results show that if the auxiliary loss (adversarial and/or reconstruction) is made too important compared to the translation loss, the translation quality is impaired. \\

The $G$-WGAN also shows improvements over the baseline and obtains similar results to $Q$-WAAE. Nonetheless, a small discrepancy is noticeable on the COCO-ambiguous. We believe that the main advantage of the $Q$-WAAE loss is the actual presence of a direct mean square error reconstruction loss along the adversarial loss.  We also noticed that the $G$-WGAN model is really sensitive to the dimension of noise concatenated to the hidden state given as input to the generator as stated in table \ref{wgannoise}. \\
\begin{table}[!h]
  \centering
\def\arraystretch{1.5}
\begin{tabular}{cc|cccc}
\multicolumn{2}{c}{}
            &   \multicolumn{4}{c}{$|z|$} \\
    &       &   64 &   128   &  256   &  512        \\ 
    \cline{2-6}
\multirow{3}{*}{\rotatebox[origin=c]{90}{}}
    & METEOR   & 50.35   &  \textbf{50.43}     & 49.71 & 49.48
    \end{tabular}
\caption{$G$-GWAN : Impact of the noise concatenated to the hidden state of size 512}
\label{wgannoise}
\end{table}

One can argue that because the generator is conditional on the hidden state $h_T$ which is of high dimension, its very hard for the generator to become deterministic. An important noise dimension could potentially harm the generator instead of fooling the discriminator.

\subsection{Quality evaluation}

To understand the success of $Q$-WAAE on the ambiguous COCO data-set, we perform an ablation study of the model. We first discard the adversarial discriminator so that we only train the reconstruction module with the MSE loss (+ $G$). We also discard the use of the features $v$ in the translation model for both the ablated model and $Q$-WAAE (no $v$). The results of the ablation study can be found in table \ref{abla}. 

\begin{table}[h]
\centering
\begin{tabular}{llllllll}
            \multicolumn{1}{c}{\bf Test sets}  &\multicolumn{2}{c}{\textbf{COCO-ambiguous}} \\
            \\ \hline \\
            &\multicolumn{1}{l}{\bf 	BLEU}   &\multicolumn{1}{l}{\bf 	METEOR} \\
            \\
            Baseline & 28.50 & 48.80  \\
            Baseline + $G$ + no $v$  & 29.43  & 49.60  \\
            Baseline + $G$  & 29.91 & 49.24  \\
            $Q$-WAAE + no $v$  & 30.57 & 50.15  \\
            $Q$-WAAE  & 31.41 & 50.95  \\
            \hline \\
\end{tabular}
\caption{Ablation study of $Q$-WAAE model}
\label{abla}
\end{table}

A first observation is that the reconstruction module $G$ does improve the baseline, but the the Baseline + $G$ + no $v$ model (no the visual input in the translation pipeline) has a better METEOR metric than the Baseline + $G$ model. It means that use of a visual attention model in the translation pipeline harms the overall translation quality, as already found in previous work. In contrast, $Q$-WAAE hopefully performs better than $Q$-WAAE + no $v$, which shows the successful  integration of the visual input, as it should be expect for the MMT task. Using adversarial feedback does provide a stronger training and a better generalization over the different data-sets.

\subsection{Improvements examples}

\begin{figure}[ht!]
	\centering
	\includegraphics[scale=0.26]{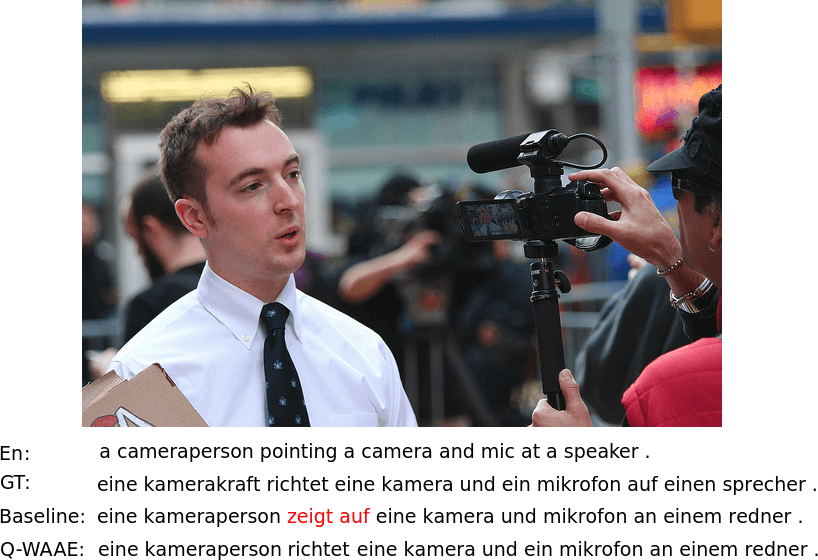}
	\caption{An ambiguous COCO example where $Q$-WAAE finds the right translation for the verb}
	    \label{jbw}
\end{figure}

To further investigate the quality of the $Q$-WAAE model, we pick two examples to illustrate the improvements. \\

In figure \ref{jbw}, the baseline translates "pointing a camera" to "zeigt auf ein camera" which could translate to "to point at a camera". It is incorrect since the image displays the camera-man pointing \textit{a} camera \textit{at} the speaker. Also, the german verb "zeigen" also means to show, to demonstrate, which is not ideal in this example. Our model translates "pointing" to "richtet" meaning "pointing" with the idea of aiming which is more suitable. Also $Q$-WAAE does not use wrong prepositions. The sentence of baseline scores a BLEU of 0 while the sentence score of our model is a BLEU of 44.83.

\begin{figure}[ht!]
	\centering
	\includegraphics[scale=0.35]{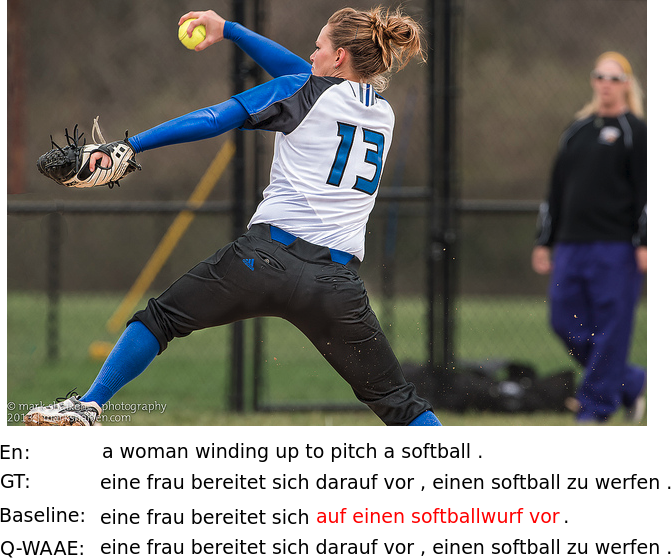}
	\caption{An ambiguous COCO example where $Q$-WAAE finds the right translation for the object}
	    \label{jbw2}
\end{figure}

The second figure aims to show that not only $Q$-WAAE manages to correctly translates ambiguous verbs but more complex examples. In Figure \ref{jbw2}, the $Q$-WAAE model ends up getting the perfect translation (a BLEU score of 100) whereas the baseline model outputs a translation closer to "a woman winding up for softball", missing the second verb (BLEU score of 22.60). 

\subsection{Other data-set}

We decided to train $Q$-WAAE on another language pair of the Multi30K dataset, namely the en $\rightarrow$ fr pair. Again the model surpasses the baseline for the COCO-ambiguous and test 2018 test sets. 

\begin{table}[t]
\centering
\begin{tabular}{lcc}
              &\multicolumn{1}{l}{\bf 	BLEU}   &\multicolumn{1}{l}{\bf 	METEOR} \\
             \hline 
            \multicolumn{1}{c}{\bf Test sets en $\rightarrow$ fr}&\multicolumn{2}{c}{\textbf{COCO-ambiguous}} \\
            \hline 
            DeepGru  & 46.16 & 65.79  \\
            $Q$-WAAE  & 47.00 & 66.50 \\ \\
            \multicolumn{1}{c}{}  &\multicolumn{2}{c}{\textbf{Test 2017}} \\
            \hline 
            DeepGru & 55.13 & 71.52  \\
            FAA  & 52.80 & 69.60  \\
            $Q$-WAAE  & 56.54 & 72.32  \\ \\
            
            \multicolumn{1}{c}{}  &\multicolumn{2}{c}{\textbf{Test 2018}}
            \\ \hline
            FAA & 39.48 & 59.85	 \\
            $Q$-WAAE  & \textbf{40.09 }& \textbf{60.54}  \\
            \hline \\
\end{tabular}
\caption{Results on the en $\rightarrow$ fr Multi30K dataset, test 2018 results can found online in the aformentioned codalab link in the "french" tab}
\end{table}

\section{Conclusion}
We demonstrated that recent advances in adversarial generative modeling was able to successfully ground visual information for multi-modal translation using visual and linguistic input. We show that the use of visual information for the model still remains a challenging task. The presented work in this paper aimed to modulate the last hidden state at the end of the translation model, it would be interesting to investigate adversarial approaches more upstream in the pipeline like in the visual features extraction (as previously investigated in \cite{delbrouck2017modulating}).

\bibliography{acl2019}
\bibliographystyle{acl_natbib}

\clearpage

\appendix
\section{Model $Q$} \label{app:modelq}
\label{sec:appendix}

Given a source sentence $\vx$ and visual features $\vv$, an attention-based encoder-decoder model outputs the translated sentence $\vy$. If we denote $\vtheta$ as the model parameters, then $\vtheta$ is learned by maximizing the likelihood of the observed sequence $\vy$ or in other words by minimizing the cross entropy loss. The objective function is given by:
\begin{equation}
\mathcal{L_Q}(\vtheta) = - \sum\limits_{t=1}^n \log p_{\vtheta}(\vy_t | \vy_{<t}, \vv, \vx) \label{eq:5}
\end{equation}

Three main components are involved: an encoder, a decoder and an attention model. \\

\textbf{Encoder} \quad  The encoder is a bidirectional-GRU that create a set of annotation $S$: 
\[
\mS = \begin{bmatrix}
\text{GRU}_{\text{forward}}(\overrightarrow{\vx})\\
\text{GRU}_{\text{backward}}(\overrightarrow{\vx})
\end{bmatrix}
\]

A word $x_t$ has an embedding of 256, each GRU is of size 512 thus annotation $\mS$ are of size 1024.\\

\textbf{Decoder} \quad The decoder is a conditional GRU (cGRU). The following equations describes a cGRU cell :

\begin{align} 
\vh_t^{\prime} =& ~ \text{GRU}_1(y_t,\vh_{t-1}) \nonumber \\
\vc_t =& ~ \text{ATT}(\vh_t^{\prime},\vv, \mS) \nonumber \\
\vh_t =& ~ \text{GRU}_1(\vh_t^{\prime},\vc_t) 
\end{align}	

where both GRU have 512 units and $\text{ATT}$ is the attention module defined hereafter :

\begin{align}
\va_t^{\prime}  =& \mW^a  \tanh(\mW^{\text{h}} \vh_t^{\prime} + \mW^{\text{s}} \mS) \\
\va_t =& \text{softmax} ( \va_t^{\prime} )  \\
\vc_t^{\prime} =& ~ \sum_{i=0}^{M-1} \va_{t_i} \vs_i \\
\vi_t =& ~  \tanh ( \mW^{\text{feat}} \vv  ) \label{eq:vt} \\
\vc_t =& ~  \mW^c (\vc_t^{\prime} \odot \vi_t)   \label{eq:vt_merge}
\end{align}

Matrices $\mW^s$ and $\mW^h$ map respective inputs to size 1024 $\mW^h$. $\mW^{\text{feat}}$ transform visual features to size 1024 and $\mW^c$ transforms both attention vector back to size 512 to be compatible with $\text{GRU}_2$ size. \\

Finally, a bottleneck function projects the cGRU output into probabilities over the target vocabulary. It is defined so:

\begin{align}
\vb_t &= \tanh(\mW^{\text{bot}}\vh_t \label{f_bt})  \\ 
y_t \sim \vp_t &= \text{softmax}(\mW^{\text{proj}} \vb_t) \label{f_proj}
\end{align}

where $\mW^{\text{bot}}$ maps hidden state to size 256 and $\mW^{\text{proj}}$ maps the bottleneck result to the vocabulary size. \\

Dropout of 0.3 is used on embeddings $\vx$ and annotations $\mS$ and of 0.5 on $\vb_t$. \\

To marginally reduce our vocabulary size, we use the byte pair encoding (BPE) algorithm on the train set to convert space-separated tokens into sub-words \cite{P16-1162}. With 10K merge operations, the resulting vocabulary sizes of each language pair are: 5204 $\rightarrow$ 7067 tokens for English$\rightarrow$ German and 5835$\rightarrow$ 6577 tokens for English$\rightarrow$French.

\section{Generator $G$ and discriminator $D$} \label{app:spectral}

\textbf{$Q$-WAAE} \quad Generator G is defined as follows: 

$$ \vv^\prime =\tanh(\mW^{\text{rec}} h_T)$$

where $\mW^{\text{rec}}$ is of size $512\times2048$. \\

Discriminator D is defined as follows : 

$$ \vo =\mW^{\text{adv}} h_T$$
where $\mW^{\text{adv}}$ is of size $512\times1$. \\

\textbf{$G$-WGAN} \quad Generator G is defined as follows: 

$$ \vv^\prime =\tanh(\mW^{\text{rec}} [z,h_T])$$

where $\mW^{\text{rec}}$ is of size $640\times2048$. \\

Discriminator D is defined as follows ($\vv$ is either real $\vv$ or generated $\vv^\prime$): 
\begin{align}
     \vo_1 &=\text{relu}(\mW^{\text{adv}_1} [\vv, h_T]) \\
      \vo_2 &=\text{relu}(\mW^{\text{adv}_2} \vo_1) \\
      \vo_3 &= \mW^{\text{adv}_3} \vo_2
\end{align}
where $\mW^{\text{adv}_1}$ is of size $2560\times1024$, $\mW^{\text{adv}_2}$ of size is of size $1024\times512$ and  $\mW^{\text{adv}_3}$ of size is of size $512\times1$

% \subsection{PyTorch Code}

% \textbf{$W$-GAN} \quad
% \begin{minted}[bgcolor=white,formatcom=\color{black},breaklines]{python}

% def compute_gradient_penalty(D, real_samples, fake_samples):
%     """Calculates the gradient penalty loss for WGAN GP"""
%     # Random weight term for interpolation between real and fake samples
%     alpha = Tensor(np.random.random((real_samples.size(0), 1)))
%     # Get random interpolation between real and fake samples
%     interpolates = alpha * real_samples + ((1 - alpha) * fake_samples)
%     interpolates = Variable(interpolates, requires_grad=True)
%     d_interpolates = D(interpolates)
%     fake = Variable(Tensor(real_samples.shape[0], 1).fill_(1.0), requires_grad=False)
%     # Get gradient w.r.t. interpolates
%     gradients = autograd.grad(
%         outputs=d_interpolates,
%         inputs=interpolates,
%         grad_outputs=fake,
%         create_graph=True,
%         retain_graph=True,
%         only_inputs=True,
%     )[0]
%     gradients = gradients.view(gradients.size(0), -1)
%     gradient_penalty = ((gradients.norm(2, dim=1) - 1) ** 2).mean()
%     return gradient_penalty
% \end{minted}

\end{document}